\documentclass[twocolumn]{article}

\usepackage{authblk}
\usepackage{graphicx}
\usepackage{diagbox}
\usepackage{caption2}
\usepackage{color}
\usepackage{cite}
\usepackage{graphicx}
\usepackage{subfigure}
\usepackage{footnote}
\makesavenoteenv{table}
\usepackage{threeparttable}
\usepackage[colorlinks,linkcolor=blue]{hyperref}

\title{Adapted Center and Scale Prediction: \\More Stable and More Accurate}
\author[1]{Wenhao Wang\thanks{Corresponding author: wangwenhao0716@gmail.com}}
\affil[1]{University of Technology Sydney}
\author[2]{Jusheng Zhang}
\affil[2]{Sun Yat-sen University}

\date{} 
\providecommand{\keywords}[1]{\textbf{Keywords:} #1}
\begin{document}
\bibliographystyle{plain}
\maketitle
\begin{abstract}
Pedestrian detection benefits from deep learning technology and gains rapid development in recent years. Most of detectors follow general object detection frame, \textit{i.e.} default boxes and two-stage process. Recently, anchor-free and one-stage detectors have been introduced into this area. However, their accuracies are unsatisfactory. Therefore, in order to enjoy the simplicity of anchor-free detectors and the accuracy of two-stage ones simultaneously, we propose some adaptations based on a detector, Center and Scale Prediction(CSP). The main contributions of our paper are: $(1)$ We improve the robustness of CSP and make it easier to train. $(2)$ We propose a novel method to predict width, namely compressing width. $(3)$ We achieve the second best performance  on CityPersons benchmark, \textit{i.e.} $9.3\%$ log-average miss rate(MR) on reasonable set, $8.7\%$ MR on partial set and $5.6\%$ MR on bare set, which shows an anchor-free and one-stage detector can still have high accuracy. $(4)$ We explore some capabilities of Switchable Normalization which are not mentioned in its original paper. The code is publicly available at https://github.com/WangWenhao0716/Adapted-Center-and-Scale-Prediction.
\end{abstract}

\keywords{Pedestrian Detection, Anchor-free, Switchable Normalization, Convolutional Neural Networks}

\section{Introduction}
With the prevalence of artificial intelligence technique, autonomous vehicles have gained more and more attention. Although automatic driving needs integration of a lot of technologies, pedestrian detection is one of the most important. That's because missing pedestrian detection could threaten pedestrians' lives. As a result, the performance of pedestrian detection algorithms is of great importance. \par 
With the development of generic object detection\cite{redmon2017yolo9000, redmon2016you, liu2016ssd, ren2015faster, girshick2014rich, girshick2015fast}, the detection performance on benchmark datasets\cite{dollar2009pedestrian, geiger2012we, zhang2017citypersons, shao2018crowdhuman, braun2019eurocity} is significant improved. Also, some detectors, such as \cite{liu2018learning, liu2019high, DBLP:journals/corr/abs-1910-09188,pang2019mask}, are specially designed for pedestrian detection. \par 
However, though detection performance is improved on benchmark datasets all the time, there is still a huge gap between current pedestrian detector and a careful people\cite{zhang2017towards}. Therefore, further performance improvement is necessary. Take pedestrian detection dataset, CityPersons\cite{zhang2017citypersons}, for instance. For a fair comparison, the following log-average miss rates(denoted as MR)(lower is better) are reported on the reasonable validation set with the same input scale (1x). From all of the state-of-the-arts literature available(including preprint ones), we summarize as follows: Repulsion Loss\cite{wang2018repulsion}($13.2\%$), OR-CNN\cite{zhang2018occlusion}($12.8\%$), HBAN\cite{lu2019semantic}($12.5\%$), ALF\cite{liu2018learning}($12.0\%$), Adaptive NMS\cite{liu2019adaptive}($11.9\%$), CSP\cite{liu2019high}($11.0\%$), MGAN\cite{pang2019mask}($10.5\%$), PSC-Net\cite{xie2020psc}($10.4\%$), APD\cite{DBLP:journals/corr/abs-1910-09188}($8.8\%$). In the aforementioned state-of-the-arts methods, most of them have special occlusion/crowd handling process($7/9$): Repulsion Loss\cite{wang2018repulsion}, OR-CNN\cite{zhang2018occlusion}, HBAN\cite{lu2019semantic}, Adaptive NMS\cite{liu2019adaptive}, MGAN\cite{pang2019mask}, PSC-Net\cite{xie2020psc}, APD\cite{DBLP:journals/corr/abs-1910-09188}. In addition, APD\cite{DBLP:journals/corr/abs-1910-09188} uses more powerful backbone, \textit{i.e.} DLA-34\cite{yu2018deep}, to improve MR from $10.6\%$(ResNet-50\cite{he2016deep}) to $8.8\%$. APD\cite{DBLP:journals/corr/abs-1910-09188} also takes advantage of post process like Adaptive NMS\cite{liu2019adaptive}.\par 
For CSP\cite{liu2019high}, there is no special occlusion/crowd handling process or more powerful backbone. And it achieves competitive MR with other methods. However, there are also some unsolved problems existing 
in CSP\cite{liu2019high}. First, it is sensitive to the batch size. More specifically, in the case of a small batch size, such as $(1,1)$(The bracket (·, ·) denotes (\#GPUs,\#samples per GPU)), or a big batch size, such as $(4,4)$, MR will not converge, \textit{i.e.} MR will approach $1$ after several iterations. Second, when training CSP\cite{liu2019high}, different input scales bring significantly different performance. Finally, when compared to algorithms with occlusion/crowd handling process, there is still much room for improvement.\par
\begin{figure*}[htbp]
\centering
\subfigure{
\includegraphics[width=7.3cm]{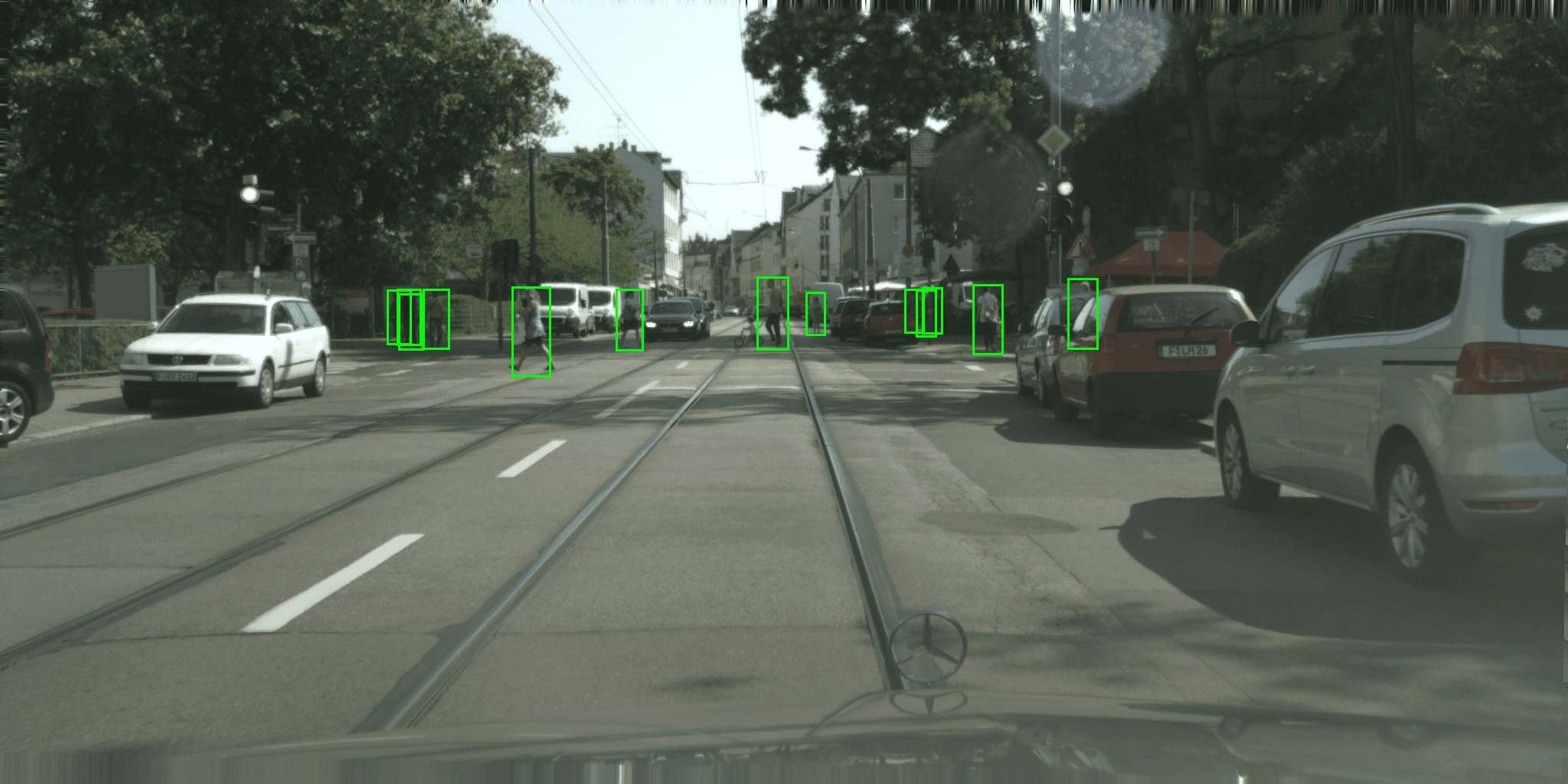}
}
\quad
\subfigure{
\includegraphics[width=7.3cm]{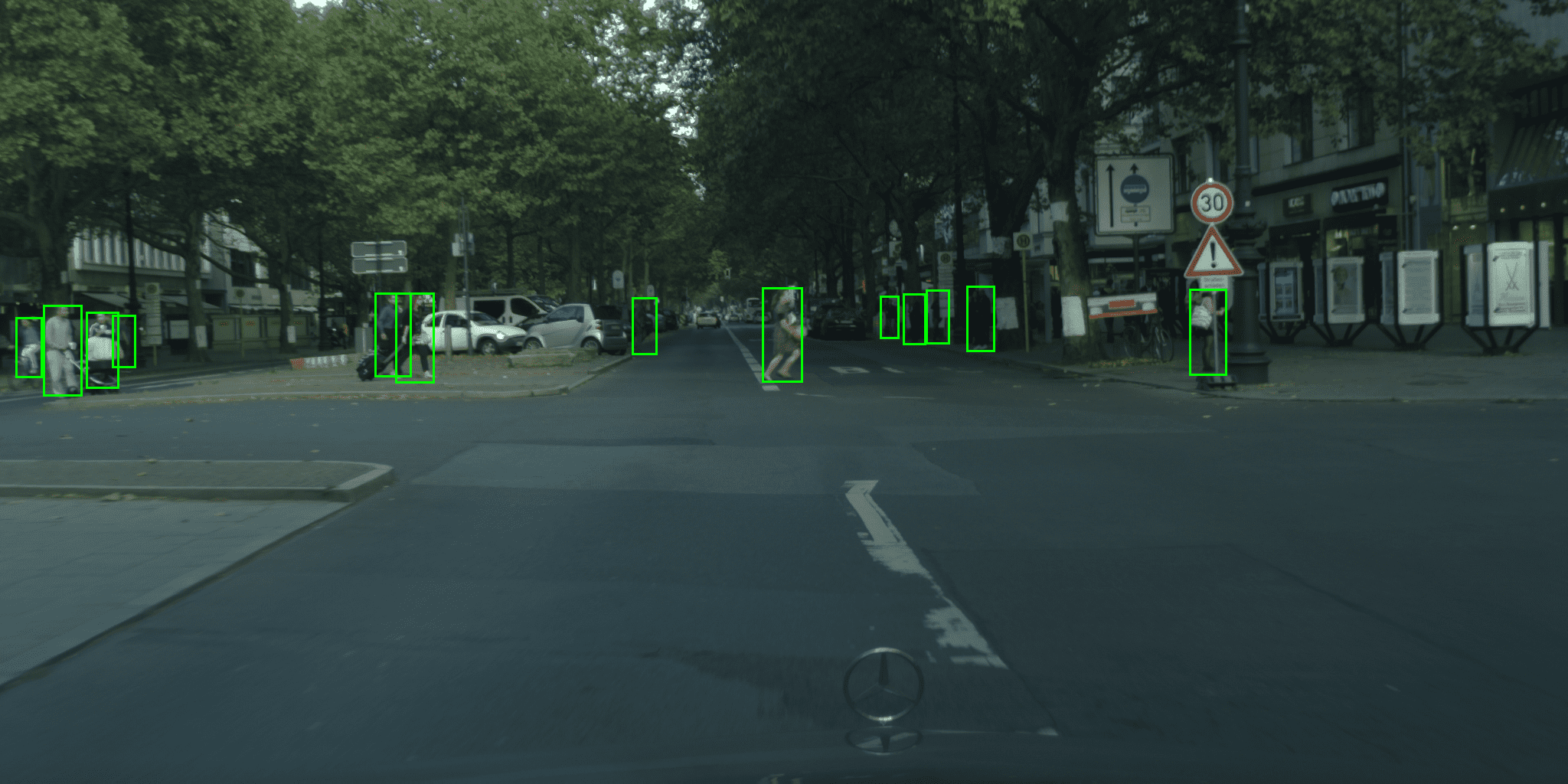}
}
\quad
\subfigure{
\includegraphics[width=7.3cm]{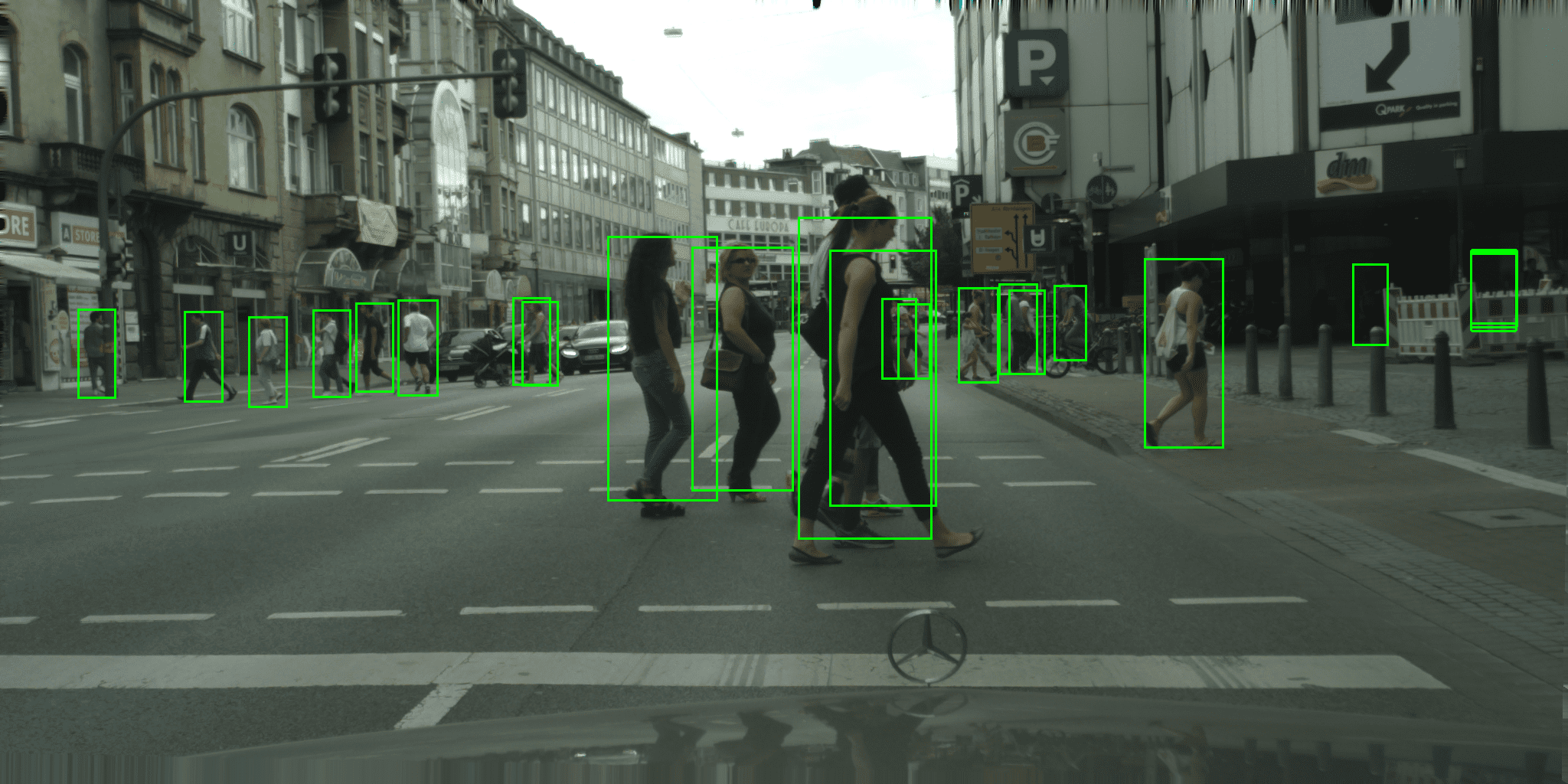}
}
\quad
\subfigure{
\includegraphics[width=7.3cm]{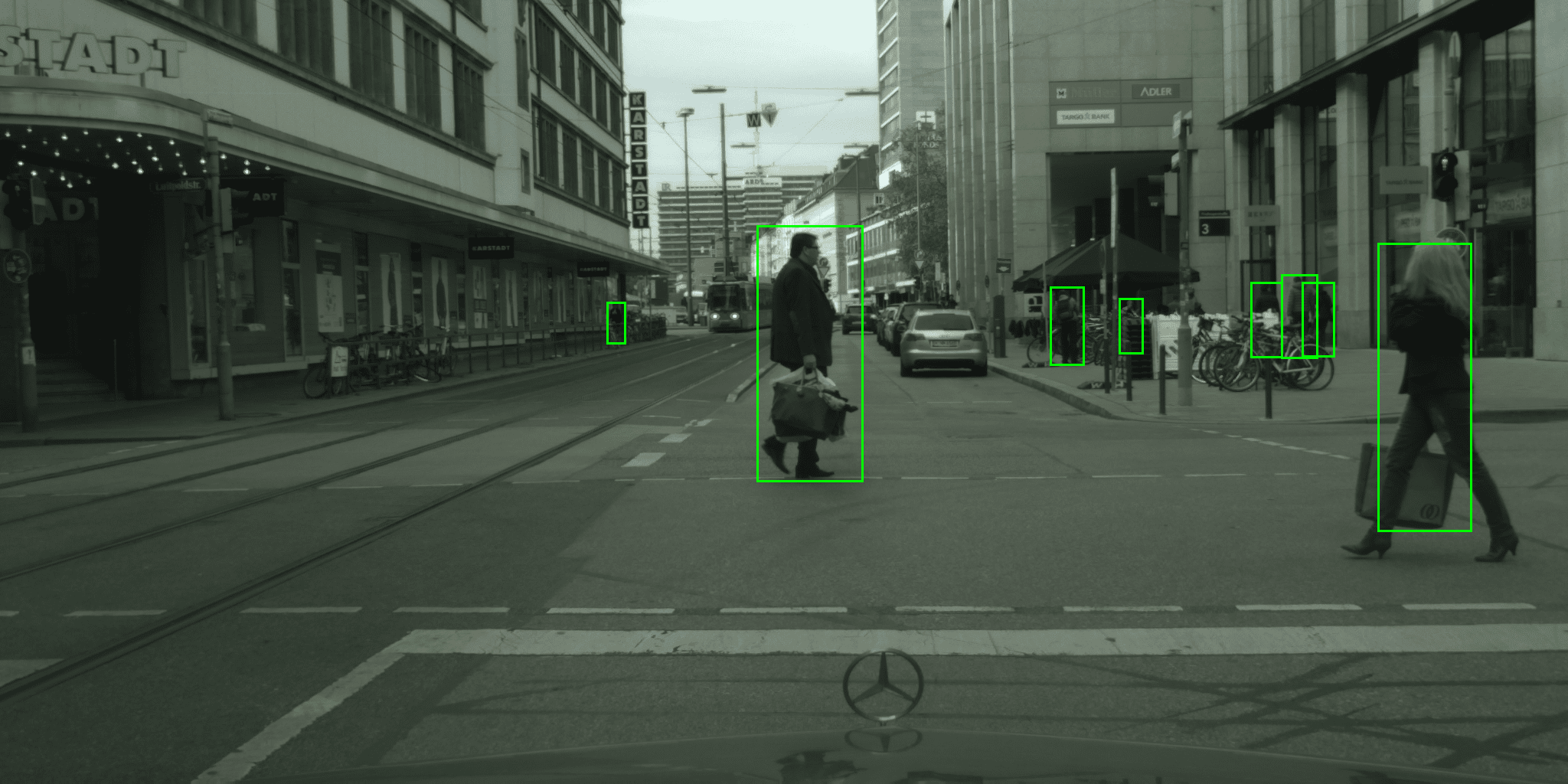}
}
\caption{We use CityPersons test set to illustrate our ACSP detection ability. It is worthy to mention that, without any occlusion handling method, small and occlusion pedestrian can still be detected. The detections are shown in green rectangle boxes.}
\label{fig0}
\end{figure*}

 
To address the above limitations, we propose \textbf{A}dapted \textbf{C}enter and \textbf{S}cale \textbf{P}rediction (\textbf{ACSP}), which has slight difference with original CSP\cite{liu2019high} but brings significant improvement on performance. Detection examples using ACSP are shown in Fig. \ref{fig0}. In summary, the main contributions of this paper are as follows: $(1)$ We make original CSP\cite{liu2019high} more robust, \textit{i.e.} less sensitive to batch size and input scale. $(2)$We propose compressing width, a novel method to determine the width of a bounding box. $(3)$We improve the performance of CSP\cite{liu2019high}.  $(4)$ We explore the power of Switchable Normalization when the batch size is big.\par 
Experiments are conducted on the CityPersons\cite{zhang2017citypersons} database. We achieve the second best performance, \textit{i.e.} $9.3\%$ MR on reasonable set, $8.7\%$ MR on partial set, $5.6\%$ MR on bare set.

\section{Related Works}
\subsection{Generic Object Detection}
Early object detection approaches, such as\cite{viola2001robust, lienhart2002extended, dollar2014fast}, mainly utilize region proposal classification and sliding window paradigm. Since August 2018, more and more works focus on anchor-free approaches. As a result, modern object detectors can be divided into two classes: anchor-based and anchor-free.
\subsubsection{Anchor-based}
Anchor-based methods includes two-stage detectors and one-stage detectors. The most famous series of two-stage detectors are RCNN\cite{girshick2014rich} and its descendants, \textit{i.e.} Fast-RCNN\cite{girshick2015fast}, Faster-RCNN\cite{ren2015faster}. They build two-stage framework, which contains object proposals and classification. For one-stage detectors, YOLOv2\cite{redmon2017yolo9000} and SSD\cite{liu2016ssd} successfully accomplish detection and classification tasks on feature maps at the same time.
\subsubsection{Anchor-free}
The earliest exploitation of anchor-free mode comes from DenseBox\cite{huang2015densebox} and YOLOv1\cite{redmon2016you}. The main difference between them is that DenseBox is designed for face detection while YOLOv1 is a generic object detection. The introduction of CornerNet\cite{law2018cornernet} brings anchor-free detection into key point era. Its followers include ExtremeNet\cite{zhou2019bottom}, CenterNet\cite{zhou2019objects}, etc. In addition, FoveaBox\cite{kong2019foveabox} and FSAF\cite{zhu2019feature} use dense object detection strategy.
\subsection{Pedestrian Detection}
Before the dominance of deep learning techniques, traditional pedestrian detectors, such as\cite{dollar2014fast, nam2014local, zhang2015filtered}, focus on integral channel features with sliding window strategy. Recently, with the introduction of Faster RCNN\cite{ren2015faster}, some two-stage pedestrian detection approaches\cite{zhang2017citypersons, pang2019mask, zhou2018bi, zhou2019discriminative, zhang2018occluded, liu2019adaptive, zhang2018occlusion, wang2018repulsion} achieve state-of-the-arts on standard benchmarks. Also, some pedestrian detectors\cite{li2018csrnet, liu2018learning, liu2019adaptive}, which base on single-stage backbone, gain a balance between speed and accuracy. \par 
Zhou \textit{et al.}\cite{zhou2019discriminative} propose a discriminative feature transformation which enforces feature separability of pedestrian and non-pedestrian examples to handle occlusions for pedestrian detection. In \cite{zhang2018occlusion}, a new occlusion-aware R-CNN is proposed to improve the detection accuracy in the crowd. Wang  \textit{et al.}\cite{wang2018repulsion} develop a novel loss, repulsion loss, to address crowd occlusion problem. The work of \cite{liu2019adaptive} focuses on Non-Maximum Suppression and proposes a dynamic suppression threshold to refine the bounding boxes given by detectors. HBAN\cite{lu2019semantic} is introduced to enhance pedestrian detection by fully utilizing the human head prior. ALFNet is proposed in \cite{liu2018learning} to use asymptotic localization fitting strategy to evolve the default anchor boxes step by step into precise detection results. MGAN\cite{pang2019mask} emphasizes on visible pedestrian regions while suppressing the occluded ones by modulating full body features. PSC-Net\cite{xie2020psc} is designed for occluded pedestrian detection. CSP\cite{liu2019high} utilizes an anchor-free method, \textit{i.e.} directly predicting pedestrian center and scale through convolutions. Based on CSP\cite{liu2019high}, Zhang \textit{et al.} propose APD\cite{DBLP:journals/corr/abs-1910-09188} to differentiate individuals in crowds. All of the aforementioned methods achieve state-of-the-arts on CityPersons benchmark\cite{zhang2017citypersons}.
\begin{figure*}[h]
\centering
\includegraphics[width=16cm]{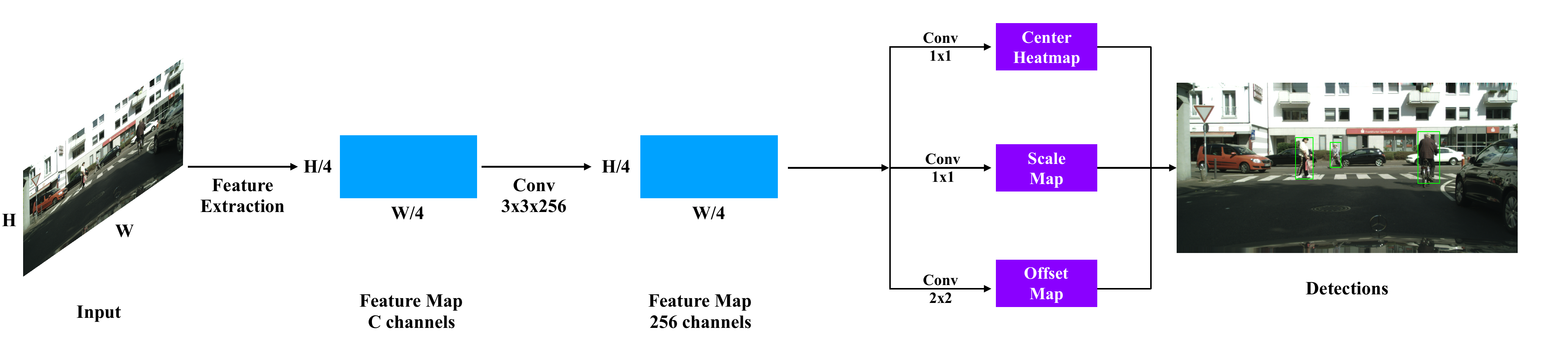}
\caption{It is the architecture of original CSP\cite{liu2019high}. The frame includes two parts: feature extraction and detection head.}
\label{fig:1}
\end{figure*}
\subsection{Normalization}\label{Norm}
Batch Normalization(BN)\cite{ioffe2015batch} is proposed to accelerate training process and improve the performance of CNNs. \cite{santurkar2018does} points out that batch normalization makes the loss surface smoother while the original paper\cite{ioffe2015batch} believes the improvement comes from "internal covariate shift". Although, even today, it is still unknown that why batch normalization works so well, the utilization of batch normalization improves the performance of object detection, image classification, etc.\par 
After batch normalization, weight normalization(WN)\cite{salimans2016weight} is introduced to normalize the weights of layers. Layer normalization(LN)\cite{ba2016layer} normalizes the inputs across the features instead of the batch dimension. In this way, the performance will not be influenced by batch size and layer normalization is used in RNN at first. Originally designed for style transfer, instance normalization(IN)\cite{ulyanov2016instance} normalizes across each channel in each training example. Group normalization(GN)\cite{wu2018group} divides the channels into groups and computes the mean and variance for normalization within each group. As a result, it addresses the problem that, when the batch size becomes smaller, the performance of batch normalization goes down. It is a combination of layer normalization and instance normalization to some degree. \par 
Recently, Luo \textit{et al.} propose switchable normalization(SN)\cite{luo2018differentiable}, which uses a weighted average of different mean and variance statistics from batch normalization, instance normalization, and layer normalization. 

\section{Proposed Adaptation}
\subsection{CSP Revisit}
CSP\cite{liu2019high} was proposed by Wei Liu and Shengcai Liao in 2019. They first introduced anchor-free method into pedestrian detection area. More specifically, CSP\cite{liu2019high} includes two parts: feature extraction and detection head. In feature extraction part, a backbone, such as ResNet-50\cite{he2016deep}, MobileNet\cite{howard2017mobilenets}, is used to extract different levels of features. Shallower feature maps can provide more precise localization information while deeper feature maps can provide high-level semantic information. In detection head part, convolutions are used to predict center, scale, and offset respectively. In Fig. \ref{fig:1}, we summarize the architecture of CSP\cite{liu2019high}.\par 

A more detailed architecture of CSP\cite{liu2019high} will be revisited in this paragraph. However, it will be slightly different with original paper\cite{liu2019high}. We take ResNet-50\cite{he2016deep} and a picture with original shape, \textit{i.e.} $1024 \times 2048$ for instance. The difference between keeping original shape and resizing picture to $640 \times 1280$ as showed in \cite{liu2019high} will be discussed in ablation study. First, CSP\cite{liu2019high} enlarges a picture with $3$ channels into $64$ channels through a 7x7 \textit{Conv} layer. Certainly, \textit{BN} layer, \textit{ReLU} layer and \textit{Maxpool} layer follow the \textit{Conv} layer. In this way, a (3, 1024, 2048)(The bracket (·, ·, ·) denotes (\#channels, height, width)) picture will be turned into a (64, 256, 512) one. Second, CSP\cite{liu2019high} take $4$ layers from ResNet-50\cite{he2016deep} with dilated convolutions. The $4$ layers downsample the input image by $4$, $8$, $16$, $16$ respectively. At that time, we get $4$ feature maps: (256, 256, 512), (512, 128, 256), (1024, 64, 128), (2048, 64, 128). CSP\cite{liu2019high} chooses to use a deconvolution layer to fuse the last $3$ multi-scale feature maps into a single one. As a result, a (768, 256, 512) final feature map is made. Third, a 3x3 \textit{Conv} layer is used on the final feature map to reduce its channel dimensions to 256. Finally, three convolutions: 1x1, 1x1 and 2x2 are appended for the prediction of center, scale and offset respectively.
\begin{figure*}[h]
\centering
\captionstyle{flushleft}
\includegraphics[width=16cm]{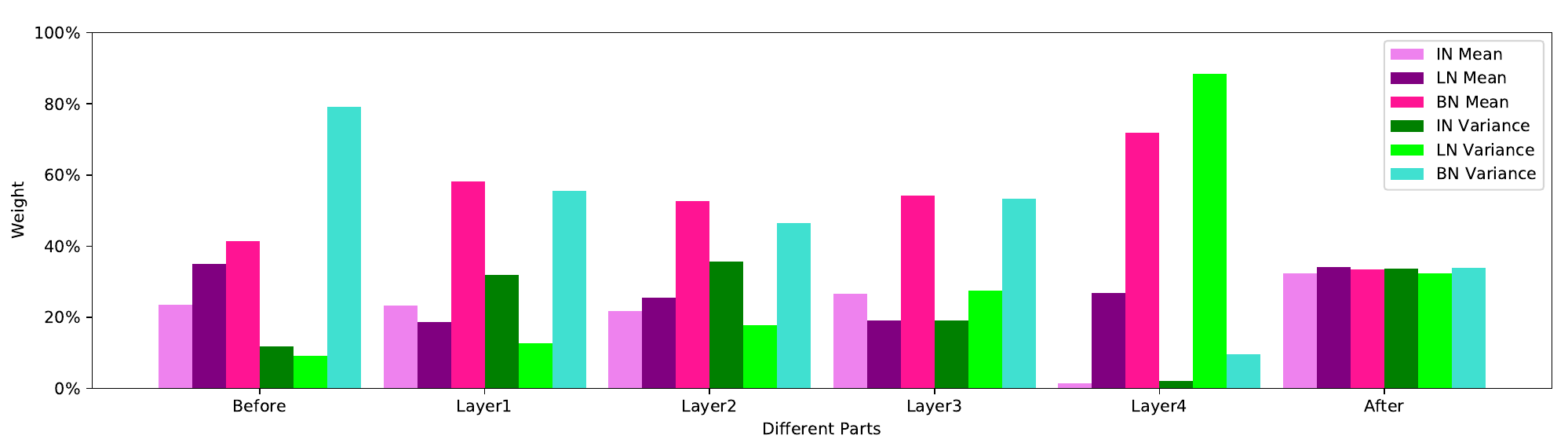}
\caption{The proportion of the weight of each normalization method in different parts is shown in the histogram. The weights of mean and variance are displayed separately.}
\label{fig:3}
\end{figure*}
\subsection{SN Layer}\label{SN}
According to the aforementioned revisit, we conclude that there are totally $50$ \textit{BN} layers in CSP\cite{liu2019high}. Although  \textit{BN} layer brings performance improvement to CSP\cite{liu2019high} as it brings to other tasks,  CSP\cite{liu2019high} also suffers from the drawback of  \textit{BN} layer. On one hand, \textit{BN} layer is unsuitable when the batch size is small. That is because small batch size will make the training process noisy, \textit{i.e.} the amplitude of training loss is relatively huge. However, ablation study will show a bigger batch size, even $(1,8)$, brings more harm to CSP\cite{liu2019high}. More specifically, MR of validation set will decrease to near $16\%$ and then increase to $1$. It is likely that CSP\cite{liu2019high} falls into local optimum and loses generalization ability.\par 
To address this limitation, we replace all \textit{BN} layers with \textit{SN} layers. The effectiveness of this change will be shown in the ablation part, we try to explain the reason of it now.\par To illustrate more specifically, we take $(1,8)$ for instance and the backbone is ResNet-50. The architecture of network can be divided into $6$ parts: The first $5$ come from backbone and the last one(denoted as After) is detection head. The first part(denoted as Before) is the operations before $4$ layers in ResNet-50. The next four is the four layers. There is only $1$ \textit{BN} layer in the first part while there are $9$, $12$, $18$, and $9$ \textit{BN} layers in the next four parts respectively. Finally, the detection head has only $1$ \textit{BN} layer. As suggested in \ref{Norm}, Switchable normalization is the combination of batch normalization, instance normalization, and layer normalization with different weights. Therefore, exploring the proportion of different weights in each part of network will show what makes a difference on earth. For each part, we calculate the weights of each normalization method in \textit{SN} layers. Then the average of these weights are shown in Fig. \ref{fig:3}. Although different normalization methods have different weights in each part, we figure out two main differences with original \textit{BN} layers. For one hand, in Layer4, the weight of BN variance is very small while the weight of LN variance is very big. For the other hand, in 'After' Part, IN, LN and BN share similar weights. The conclusions are as follows: $(i)$ The low BN variance in Layer4 decreases the influence of noise when estimating variance. In this way, high-level semantic information can be utilized fully during inference process. $(ii)$ The similar weights in 'After' part enable these three normalization methods to play same important roles. $(iii)$ Different normalization methods in all parts complement one another.

\subsection{Backbone}\label{BB}
The feature extraction ability of backbone is of great importance in object detection. Some networks, such as, ResNet-50\cite{luo2018differentiable}, ResNet-101\cite{luo2018differentiable},  VGG\cite{simonyan2014very} and MobileNet\cite{howard2017mobilenets}, which are original designed for image classification, are 
widely used in pedestrian detectors. In addition, some other networks, such as DetNet\cite{li2018detnet}, are specially designed for object detection.\par 
In the original paper\cite{liu2019high}, ResNet-50\cite{luo2018differentiable} and MobileNet\cite{howard2017mobilenets} are used as backbone. However, because of the nature of CSP\cite{liu2019high}, \textit{i.e.} it fuses different level of feature maps, it is suitable to use deeper backbone network. In this way, the location information will still be stored  in shallow feature maps and higher-level semantic information will be extracted at the same time.\par 
Inspiring by the aforementioned idea, we select two new backbones, expecting to obtain better performance. First, we use ResNet-101\cite{luo2018differentiable} as our ACSP backbone. Compared to ResNet-50\cite{luo2018differentiable}, the only difference of ResNet-101\cite{luo2018differentiable} is its third layer: there are $23$ Bottleneck blocks rather than $6$ Bottleneck blocks. As a result, in our ACSP, the last two feature maps presents higher level semantic information than CSP\cite{liu2019high}. Meanwhile, localization information will not be changed. In theory, the fusion in our ACSP is more efficient than original CSP\cite{liu2019high}. We will conduct ablation study to prove it. Second, in \cite{li2018detnet}, authors point out that using DetNet\cite{li2018detnet} as backbone, they achieve state-of-the-art on the MSCOCO benchmark\cite{lin2014microsoft}. Therefore, it is likely that DetNet\cite{li2018detnet} will improve the performance of original CSP\cite{liu2019high}. However, after fine tuning learning rate and so on, we find it is unpromising. We conclude the reason is that: one of the design concept of DetNet\cite{li2018detnet} is to address poor location problem, however, in CSP\cite{liu2019high}, this problem is solved by the fusion of different level layers and efficient center prediction.

\subsection{Input Size}\label{is}
In the original paper\cite{liu2019high}, Liu \textit{et al.} do not justify the resizing process, \textit{i.e.} why in training part, the authors resize the original picture shape($1024 \times 2048$) into $640 \times 1280$. After comparison, we find that: For one hand, resizing shape is beneficial to time-saving and memory-saving. More importantly, keeping original shape will worsen the performance of CSP\cite{liu2019high} and bring non-convergent results. However, most of its counterparts take advantage of original resolution and achieve state-of-the-arts. Inspiring by this, we believe some adjustment will make a difference.\par 
Based on the improvement in \ref{SN}, we compare the performance between keeping and resizing. In ablation part, we will show that the performance also becomes worse, however. We conclude the reasons: First, some noise may exist in the original pictures. With the ResNet-50 backbone, the semantic features are not extracted adequately. Therefore, parameters of the network may be influenced by noise. Second, the quanlity of parameter is not sufficient to fit the useful part of so high resolution pictures. Finally, resizing process will omit some detail features, and focusing on them excessively will influence the ability of generalization.\par 
To address the aforementioned problems, we replace ResNet-50 with ResNet-101 as suggested in \ref{BB}. In this way, the performance is improved and we achieve the lowest MR of our ACSP. The reasons are as follows: For one hand, the increase of parameters enhances fitting ability of our ACSP. For the other hand, the increase of layers enables our ACSP to extract more high-level semantic feature and decrease the focus on details. Therefore, ACSP is immune to noise and has more generalization ability.

\subsection{Compressing Width}
From the original paper\cite{liu2019high}, we can see that the width of a box is obtained by multiplying the height by $0.41$. It concurs with pedestrian aspect ratio in CityPersons Dataset\cite{zhang2017citypersons}. However, it is not suitable in the reference process. That is because, in crowded scene, relatively wide boxes will increase the chance of overlapping and the NMS process will eliminate some of boxes. In this way, we will lose some detections.\par 
As a result, we try to design a novel method to determine the width. On one hand, as we mentioned before, a wide box is not appropriate. On the other hand, a too narrow box is also not suitable. That is because, in this way, IoU between detections and ground truths will be small and detections will not be regarded as correct. 
Inspiring by the aforementioned analysis, we give our formula for calculating width:
\[w = r \cdot h,\]
where $r$ is the aspect ratio($r<0.41$) and $h$ is the predicted height of a bounding box.\par
It should be mentioned that the exact form of our compressing width is not crucial and we choose the most basic one. What matters most is the design concept. 

\subsection{Vanilla L1 Loss}
As pointed in \cite{liu2019high}, total loss consists of classification loss, scale loss and offset loss. The weights are 0.01, 1 and 0.1, respectively. And for scale regression loss, \cite{liu2019high} utilizes Smooth L1 to accelerate convergence.  However, \cite{zhou2019objects,sun2017compositional,sun2018integral} show that vanilla L1 is better than Smooth L1. Therefore, we try to replace Smooth L1 with vanilla L1. We experimentally set the weights as 0.01, 0.05 and 0.1, respectively. The effectiveness of this improvement will be shown in ablation study.
\begin{table*}[htbp] 
\centering
\captionstyle{flushleft}
\setlength{\belowcaptionskip}{10pt} 
\caption{Comparisons of different batch sizes and different methods. The bracket (·, ·) denotes (\#GPUs,\#samples per GPU). 'Con' means the training is convergent while 'Exp' means the training is not convergent. Bold number indicates the best result.}
\setlength{\tabcolsep}{3pt}
\begin{tabular}{|c|c|c|c|c|c|c|}
\hline
\diagbox{method}{MR}{batch}&
$(4, 2)$& 
$(4, 4)$& 
$(2, 2)$&
$(1, 1)$& 
$(1, 8)$&
$(8, 1)$\\
\hline
CSP&$11.56\%$ Con&$27.75\%$ Exp&$11.34\%$ Con&$16.35\%$ Exp&$16.10\%$ Exp&$14.51\%$ Exp\\
\hline
ACSP&$11.16\%$ Con&$11.89\%$ Con&$\textbf{10.80\%}$ Con&$13.42\%$ Con&$11.66\%$ Con&$12.88\%$ Con\\
\hline
Improvement&$+3.46\%$&$+57.15\%$&$+4.76\%$&$+17.92\%$&$+27.58\%$&$+11.23\%$ Con\\
\hline
\end{tabular}
\label{tab1}
\end{table*}

\section{Experiments}
\subsection{Experiment settings}
\subsubsection{Dataset}
To prove the efficacy of our adaptation, we conduct our experiments on CityPersons Dataset\cite{zhang2017citypersons}. CityPersons is introduced recently and with high resolution. And the dataset is based on CityScapes benchmark\cite{cordts2016cityscapes}. It includes $5,000$ images with various occlusion levels. We train our model on official training set with $2,975$ images and test on the validation set with $500$ images. In our test, the input scale is 1x.
\subsubsection{Training details}
The ACSP is realized in Pytorch\cite{paszke2017automatic}. Adam\cite{kingma2014adam} optimizer is utilized to optimize our network. Same as CSP\cite{liu2019high} and APD\cite{DBLP:journals/corr/abs-1910-09188}, moving average weights\cite{tarvainen2017mean} is adopted. Experiments show it helps achieve better performance. The backbone is fixed ResNet-101\cite{luo2018differentiable} unless otherwise stated, \textit{i.e.} replacing all \textit{BN} layers with \textit{SN} layers. It is pretrained on ImageNet\cite{deng2009imagenet}. We optimize the network on 2 GPUs (Tesla V100) with $2$ images per GPU. The learning rate is $2 \times 10^{-4}$ and training process is stopped after $150$ epochs with 744 iterations per epoch. In the training process, we keep the original shape of pictures, \textit{i.e.} $1024 \times 2048$. 

\subsection{Ablation Study}
In this section, we conduct an ablative analysis on the CityPersons Dataset\cite{zhang2017citypersons}. We use the most common and standard pedestrian detection criterion, log-average miss rates(denoted as MR), as evaluation metric. In addition, the following MRs are all reported on reasonable set. \par 
\textbf{What is the influence of \textit{SN} layer on stable training?}\par
The stability of training process is of great importance. It comes from two aspects: whether the network is sensitive to the batch size and whether the performance will become poor after many iterations. To answer these two questions, we compare our ACSP with original CSP\cite{liu2019high}. It should be mentioned that learning rate is  appropriate in the following experiments,  \textit{i.e.} the training loss decreases and converges.\par  
For the first one, comparisons are shown in Table \ref{tab1}. To conduct a fair comparison, the only difference is we replace all \textit{BN} layers with \textit{SN} layers, \textit{i.e.} the backbone is still Resnet-50, the training input scale is still $640 \times 1280$  and so on. In the table, the bracket (·, ·) denotes (\#GPUs,\#samples per GPU). For instance, $(4, 4)$ means $4$ GPUs with $4$ images per GPU. 'Con' means the training is convergent, \textit{i.e.} MR is still low no matter how many iterations are used. 'Exp' means the training is not convergent, \textit{i.e.} MR increases to $1$ after several iterations. The improvement line shows the percentage of decrease in MR from CSP\cite{liu2019high} to ACSP. It is shown that when we choose GPU number and image number per GPU carefully, such as $(4, 2)$, $(2, 2)$, although ACSP outperforms CSP\cite{liu2019high} to some degree, the improvement is not significant. However, when the batch size is bigger or smaller, such as  $(4, 4)$, $(1, 1)$, ACSP brings conspicuous improvement. It is noteworthy that, though batch size is $8$, there is a huge difference in MR between $(4, 2)$ and $(1, 8)$ for CSP\cite{liu2019high}. That difference does not come from \textit{BN} layer because  \textit{BN} layer will only be invalid when it is $(8, 1)$ rather than $(1, 8)$. Therefore, it is impossible to reproduce the reported result in \cite{liu2019high} for someone who only has single GPU resource. \par
\begin{figure}[h]
\centering
\includegraphics[width=8cm]{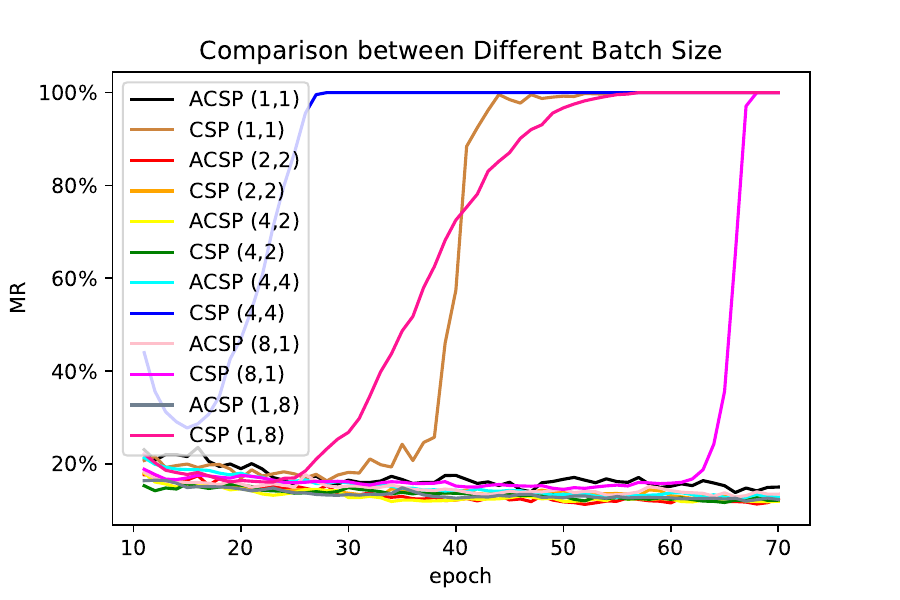}
\caption{Comparisons of different batch size. It is shown that: For CSP\cite{liu2019high}, $4$ experiment settings are not convergent; for ACSP, all experiments are convergent.}
\label{fig:4}
\end{figure}

 
For the second one, we can come to a conclusion from Figure \ref{fig:4} and Table \ref{tab1}. For CSP\cite{liu2019high}, only $(4, 2)$ and $(2, 2)$ bring convergence result. However, for ACSP, all of the results are convergent.

\textbf{How important is the backbone?}\par 
In this part, we compare three different backbones, \textit{i.e.} ResNet-50\cite{luo2018differentiable}, ResNet-101\cite{luo2018differentiable}, DetNet\cite{li2018detnet}. The experiments are conducted based on \textit{BN} layer and \textit{SN} layer respectively. The experiments setting is $(4, 2)$. And the input size is $640 \times 1280$. The results are reported in Table \ref{tab2}.\par 
We can conclude that: $(i)$ As suggested in the theory part, ResNet-101\cite{luo2018differentiable} outperforms ResNet-50\cite{luo2018differentiable} no matter which normalization method is choosen. $(ii)$ DetNet\cite{li2018detnet} underperforms ResNet-50\cite{luo2018differentiable} and ResNet-101\cite{luo2018differentiable} slightly. $(iii)$ As discussed before, replacing $BN$ layers with $SN$ layers bring performance improvement on ResNet-50\cite{luo2018differentiable} and ResNet-101\cite{luo2018differentiable}. However, on DetNet\cite{li2018detnet}, the MR increases. That is partly because we cannot find pretrained parameters of  \textit{SN} layers in DetNet\cite{li2018detnet}.

\textbf{How important is the input resolution?}\par 
To prove the discussion in \ref{is}, we conduct some experiments with different resolutions under different circumstances. From Table \ref{tab3}, we can find that: For original CSP\cite{liu2019high}, the MR is not convergent when we do not resize pictures to $640 \times 1280$. When we use SN, as expected, the MR is convergent and performance is improved. But keeping original resolution is still not a better choice.\par 
In Table \ref{tab4}, the experiments are conducted using SN as normalization method and ResNet-101 as backbone. As analysed in \ref{is}, the performance gets better no matter which batch size is chosen.

\begin{table}[h] 
\centering
\captionstyle{flushleft}
\setlength{\belowcaptionskip}{10pt} 
\caption{Comparisons of different backbones and different normalization methods. Bold number indicates the best result. The experiments setting is $(4, 2)$.}
\setlength{\tabcolsep}{3pt}
\begin{tabular}{|c|c|c|}
\hline
\diagbox{Backbone}{MR}{Method}&
BN& 
SN \\
\hline
ResNet-50&$11.56\%$ &$11.16\%$\\
\hline
ResNet-101&$11.29\%$ &$\textbf{10.91\%}$\\
\hline
DetNet&$12.66\%$&$12.91\%$\\
\hline
\end{tabular}
\label{tab2}
\end{table}
\begin{table}[htbp] 
\centering
\captionstyle{flushleft}
\setlength{\belowcaptionskip}{10pt} 
\caption{Comparison between different resolutions under different normalization methods. Resolution part means the input picture scale. The experiments setting is $(2, 2)$ and the backbone is ResNet-50.}
\setlength{\tabcolsep}{3pt}
\begin{tabular}{|c|c|c|}
\hline
\diagbox{Method}{MR}{Resolution}&
$1024 \times 2048$& 
$640 \times 1280$ \\
\hline
BN&$30.08\%$ &$11.34\%$\\
\hline
SN&$11.41\%$ &$\textbf{10.80\%}$\\
\hline
\end{tabular}
\label{tab3}
\end{table}

\begin{table}[htbp] 
\centering
\captionstyle{flushleft}
\setlength{\belowcaptionskip}{10pt} 
\caption{Comparison between different resolutions under different batch sizes. Resolution part means the input picture scale. The normalization method is SN and the backbone is ResNet-101.}
\setlength{\tabcolsep}{3pt}
\begin{tabular}{|c|c|c|}
\hline
\diagbox{Batch}{MR}{Resolution}&
$1024 \times 2048$& 
$640 \times 1280$ \\
\hline
$(2, 2)$&$\textbf{10.30\%}$&$10.81\%$\\
\hline
$(4, 2)$&$10.69\%$ &$10.91\%$\\
\hline
\end{tabular}
\label{tab4}
\end{table}

\begin{table}[htbp] 
\centering
\captionstyle{flushleft}
\setlength{\belowcaptionskip}{10pt} 
\caption{Comparisons between different aspect ratio under different sets.}
\setlength{\tabcolsep}{4pt}
\begin{tabular}{|c|c|c|c|c|}
\hline
\diagbox{Ratio}{MR}{Set}&
Reasonable& 
Heavy&Partial&Bare \\
\hline
$r=0.41$&$10.30\%$ &$46.12\%$&$9.15\%$&$6.79\%$\\
\hline
$r=0.40$&$10.00\%$ &$46.11\%$&$8.80\%$&$6.65\%$\\
\hline
\end{tabular}
\label{tab6}
\end{table}
\begin{table}[htbp] 
\centering
\setlength{\belowcaptionskip}{10pt} 
\captionstyle{flushleft}
\caption{Comparisons between different L1 loss under different sets.}
\setlength{\tabcolsep}{3.5pt}
\begin{tabular}{|c|c|c|c|c|}
\hline
\diagbox{L1}{MR}{Set}&
Reasonable& 
Heavy&Partial&Bare \\
\hline
Smooth&$10.00\%$ &$46.11\%$&$8.80\%$&$6.65\%$\\
\hline
Vanilla&$9.27\%$ &$46.34\%$&$8.66\%$&$5.62\%$\\
\hline
\end{tabular}
\label{tab7}
\end{table}

\begin{table*}[h] 
\centering
\begin{threeparttable}[b]
\captionstyle{flushleft}
\setlength{\belowcaptionskip}{10pt} 
\caption{Comparisons with state-of-the-arts on validation set: The evaluation metric is MR and the input scale is 1x. The top three results are highlighted in red, green and blue, respectively.}
\setlength{\tabcolsep}{16pt}{
\begin{tabular}{|c|c|c|ccc|c|}
\hline
Method&
Backbone& 
Reasonable& 
Heavy&
Partial& 
Bare\\
\hline
FRCNN\cite{zhang2017citypersons}&VGG-16&$15.4\%$&-&-&-\\
\hline
FRCNN+Seg\cite{zhang2017citypersons}&VGG-16&$14.8\%$&-&-&-\\
\hline
TLL\cite{song2018small}&ResNet-50&$15.5\%$&$53.6\%$&$17.2\%$&$10.0\%$\\
\hline
TLL+MRF\cite{song2018small}&ResNet-50&$14.4\%$&$52.0\%$&$15.9\%$&$9.2\%$\\
\hline
Repulsion Loss\cite{wang2018repulsion}&ResNet-50&$13.2\%$&$56.9\%$&$16.8\%$&$7.6\%$\\
\hline
OR-CNN\cite{zhang2018occlusion}&VGG-16&$12.8\%$&$55.7\%$&$15.3\%$ &$6.7\%$\\
\hline
HBAN\cite{lu2019semantic}&VGG-16&$12.5\%$&$48.1\%$&-&-\\
\hline
ALF\cite{liu2018learning}&ResNet-50&$12.0\%$&$51.9\%$&$11.4\%$&$8.4\%$\\
\hline
Adaptive NMS\cite{liu2019adaptive}&ResNet-50&$11.9\%$&$54.0\%$&$11.4\%$&\textcolor[rgb]{0,0,1}{$6.2\%$}\\
\hline
CSP\cite{liu2019high}&ResNet-50&${11.0}\%$&$49.3\%$&$10.4\%$&$7.3\%$\\
\hline
MGAN\cite{pang2019mask}&VGG-16&$10.5\%$&$47.2\%$&-&-\\
\hline
PSC-Net\cite{xie2020psc}&VGG-16&$10.4\%$&\textcolor[rgb]{1,0,0}{$39.7\%$}&-&-\\
\hline
APD\cite{DBLP:journals/corr/abs-1910-09188}&ResNet-50&$10.6\%$&$49.8\%$&$9.5\%$&$7.1\%$\\
\hline
APD\cite{DBLP:journals/corr/abs-1910-09188}&DLA-34&\textcolor[rgb]{1,0,0}{$8.8\%$}&$46.6\%$&\textcolor[rgb]{1,0,0}{$8.3\%$}&\textcolor[rgb]{0,1,0}{$5.8\%$}\\
\hline
ACSP(Smooth L1)&ResNet-101&\textcolor[rgb]{0,0,1}{$10.0\%$}&\textcolor[rgb]{0,1,0}{$46.1\%$}&\textcolor[rgb]{0,0,1}{$8.8\%$}&$6.7\%$\\
\hline
ACSP(Vanilla L1)&ResNet-101&\textcolor[rgb]{0,1,0}{$9.3\%$}&\textcolor[rgb]{0,0,1}{$46.3\%$}&\textcolor[rgb]{0,1,0}{$8.7\%$}&\textcolor[rgb]{1,0,0}{$5.6\%$}\\
\hline 
\end{tabular}}
\label{tab5}
\end{threeparttable}
\end{table*}
\textbf{What is the contribution of \textit{SN} layer to the MR?}\par
As stated in the before part, \textit{SN} layer brings significant improvement when batch size is not carefully selected. In addition, from Table \ref{tab2}, we conclude \textit{SN} layer brings approximately $0.4\%$ improvement with regard to MR. Table \ref{tab3} shows no matter which solution we select, \textit{SN} layer always contributes to performance improvement. Finally, as displayed in Table \ref{tab4}, we obtain our best performance under the help of  \textit{SN} layer. In conclusion, \textit{SN} layer can substitute \textit{BN} layer totally in our ACSP.

\textbf{How important is the compressing width and vanilla L1 loss?}\par 
We talk about the contribution of the compressing width and vanilla L1 loss together in this part. Experiments show that, for Smooth L1, setting $r$ in compressing width formula as $0.40$ yields relatively good performance. And for vanilla L1 loss, $r=0.36$ is suitable. It should be mentioned that other settings may yield better results, but we choose to keep these settings in the following paragraphs(except where noted).\par 
First, we only replace $r=0.41$ with $r=0.40$, and the results are shown in Table \ref{tab6}. It can be seen that MR decreases about $0.3\%$.\par 
Second, we compare the performance of Smooth L1 with vanilla L1 under respective optimized $r$. As displayed in Table \ref{tab7}, MR decreases to varying degrees on reasonable set, partial set, and bare set. However, MR increases about $0.2\%$ on heavy set. \par

\subsection{Comparison with the State of the Arts}
We compare our ACSP with all existing state-of-the-art detectors(including preprint ones) on the validation set of CityPersons. The results are shown in Table \ref{tab5}. The evaluation metric is MR. To conduct a fair comparison, all methods are trained on the training set without any extra data(except ImageNet). When testing, the input scale is 1x. The top three results are highlighted in red, green and blue, respectively. Because the difference in training and test environment, \textit{i.e.} most of other methods use Nvidia GTX 1080Ti GPU while we use Nvidia Tesla V100 GPU, time comparing is meaningless. As a result, it will not be reported in our table.\par 
From the table, we can figure out that our ACSP achieves state-of-the-art on bare set and the second best performance on reasonable set, heavy set and partial set. On reasonable set, the best one, APD\cite{DBLP:journals/corr/abs-1910-09188}, uses more powerful backbone and other post process method. Without DLA-34, its MR will increase to $10.6\%$ instead. On heavy set, without any special occlusion handling process, we outperform other special designed methods except for PSC-Net\cite{xie2020psc}. Also, we only lags behind APD\cite{DBLP:journals/corr/abs-1910-09188} on partial set.

\section{Conclusion}
In this paper, we propose several improvements on original pedestrian detector CSP\cite{liu2019high}. In this way, the training process of our ACSP is more robust. And we try to explain why we make these adaptations and why they make a difference. What's more, we propose a novel method to estimate the width of a bounding box. In addition, we explore some functions of Switchable Normalization which are not mentioned in its original paper\cite{luo2018differentiable}. Experiments are conducted  on the CityPersons\cite{zhang2017citypersons} and we achieve state-of-the-art on bare set and  the second best performance on reasonable set, heavy set and partial set. In the future, it is interesting to explore the "representative point" rather than the "center point" of pedestrian.

\bibliography{mylib}
\end{document}